\documentclass[12pt,journal,compsoc,onecolumn]{IEEEtran}
\usepackage{amsmath,epsfig}
 \usepackage{lingmacros}
 \usepackage{tree-dvips}
\usepackage{amsmath} 
\usepackage{amssymb}
\usepackage{color}
\usepackage{pifont}
\usepackage[noadjust]{cite}

\usepackage{multirow}
\newcommand{\cmark}{\ding{51}}
\newcommand{\xmark}{\ding{55}}

\newlength{\bibitemsep}
\newlength{\bibparskip}
\let\oldthebibliography\thebibliography
\renewcommand\thebibliography[1]{
  \oldthebibliography{#1}
  \setlength{\parskip}{\bibitemsep}
  \setlength{\itemsep}{\bibparskip}
}

\def\w{{\mathbf w}} 
 
\def\V{{\mathbf V}}

\def\W{{\bf W}}
\def\WW{{\bf W}}

\def\G{{\cal G}}
\def\V{{\cal V}}
\def\E{{\cal E}}

\usepackage[ruled,vlined]{algorithm2e}

\title{One-Shot  Multi-Rate  Pruning of Graph Convolutional Networks} 
\author{Hichem Sahbi \\
$ $ \\
Sorbonne University, CNRS, LIP6,  F-75005, Paris, France 
 }

\begin{document}
 \maketitle
\begin{abstract}
In this paper, we devise a novel lightweight Graph Convolutional Network (GCN) design dubbed as Multi-Rate Magnitude Pruning (MRMP) that jointly trains network topology and weights. Our method is variational and proceeds by aligning the weight distribution of the learned networks with an a priori distribution. In the one hand,  this allows implementing any fixed pruning rate, and also enhancing the generalization performances of the designed lightweight GCNs.  In the other hand,  MRMP achieves a joint training of multiple GCNs,  on top of shared weights, in order to extrapolate accurate networks at any targeted pruning rate without  retraining their weights.  Extensive experiments conducted on the challenging task of skeleton-based recognition show a substantial gain of our lightweight GCNs particularly at very high pruning regimes.
\end{abstract}

\section{Introduction}
\label{sec:intro}
  With the re-emergence  of deep neural networks~\cite{Krizhevsky2012}, many computer vision   tasks have been successfully revisited~\cite{He2016,He2017,HuangCVPR2017,Ronneberger2015,jiu2017nonlinear,jiu2019deep}. These tasks have been handled  with more and more  accurate but {\it oversized} networks, and this makes their deployment on edge  devices very challenging.   Particularly,  in hand-gesture recognition and  human computer interaction,  edge devices are endowed with limited computational resources.  Therefore,  fast and lightweight models with high recognition performances are vital for skeleton-based  recognition.  Recent learning models applying deep networks have shown saturated recognition accuracy without substantial improvement, while  computational efficiency still remains a serious issue.  Among these  learning models, graph convolutional networks (GCNs) are known to be effective particularly on non-euclidean domains such as skeleton data~\cite{Zhua2016,zhang2020}.  At least two categories of GCNs are known in the literature; spatial and spectral.  Spectral methods first map  graph signals from the input to the Fourier domain prior to achieve convolution,  and then map back the convolved signals in the input domain \cite{kipf17,Levie2018,Li2018,Bresson16,Bruna2013,Henaff2015,spectral1997,sahbi2021learningb,mazari2019mlgcn}. Spatial methods proceed  by aggregating node signals using multi-head attention (MHA)  before applying  convolutions on the resulting node aggregates~ \cite{Gori2005,Micheli2009,Wu2019,Hamilton2017,attention2019,sahbi2010context,sahbi2021learning,sahbi2021kernel}. Spatial GCNs are known to be  more effective compared to spectral ones,  however,  their main downside resides in their  computational complexity.  Hence, a major problem  is how to make these networks lightweight   while maintaining their high accuracy \cite{DBLP:conf/cvpr/HuangLMW18,DBLP:conf/cvpr/SandlerHZZC18,DBLP:journals/corr/HowardZCKWWAA17,DBLP:conf/icml/TanL19,cai2019once,he2018soft,he2018amc,sahbi2021lightweight,sahbi2023phase}.  \\  
  
  \indent Many existing works consider  the challenge  of lightweight network design such as  tensor decomposition~\cite{howard2019}, quantization~\cite{DBLP:journals/corr/HanMD15}, distillation~\cite{DBLP:journals/corr/HintonVD15,DBLP:conf/aaai/MirzadehFLLMG20,DBLP:conf/cvpr/ZhangXHL18,DBLP:conf/cvpr/AhnHDLD19,sahbi2006hierarchy} and pruning~\cite{DBLP:conf/nips/CunDS89,DBLP:conf/nips/HassibiS92,DBLP:conf/nips/HanPTD15,sahbi2022topologically}.   Pruning methods,  in particular,   are highly effective,  and their  principle is to remove connections whose impact on the classification performances is the least important.   Two major categories  of pruning approaches  exist in the sota; structured~\cite{DBLP:conf/iclr/0022KDSG17,DBLP:conf/iccv/LiuLSHYZ17} and unstructured~\cite{DBLP:conf/nips/HanPTD15,DBLP:journals/corr/HanMD15}. The former aims at zeroing-out weights of entire filters or channels whilst the latter aims at removing  weights separately.    While  structured  methods provide  computationally more efficient networks, they  are less precise   compared to unstructured techniques; indeed, the latter produce  more flexible (and hence  more accurate) networks which are  computationally still efficient.  Magnitude pruning (MP) \cite{DBLP:journals/corr/HanMD15} is one of the widely used  unstructured methods that proceeds by removing the smallest weight connections before retraining  the resulting pruned network.  While  being able to reach any fixed (targeted) pruning rate,  MP is clearly not optimal as it {\it decouples} the training of network weights from topology.   Hence,   removed connections  cannot be recovered when retraining the pruned networks,   and this  leads   to suboptimal  performances.  Moreover,   the full retraining of the pruned networks (at multiple pruning rates) makes MP highly intractable. \\
  
  \indent  In this paper, we introduce  a novel alternative for magnitude pruning referred to as MRMP (Multi-Rate Magnitude Pruning) that allows (i) {\it coupling} the training of network weights and topology,  (ii) {\it  learning simultaneously multiple} network instances for different pruning rates,  and (iii) {\it extrapolating} accurate networks  at any unseen pruning rate without retraining.  The proposed method constrains the distribution of the learned weights to match a fixed targeted distribution  and this allows, {\it via a band-stop  mechanism}, to dropout all the connections up to a given targeted pruning rate.  The advantage of the proposed contribution is twofold; in the one hand,  it constrains the learned weights to fit a targeted distribution and this leads to better generalization. In the other hand,  it allows obtaining fully trained networks at any unseen pruning rate instantaneously without weight retraining.  \\
  
Considering all the aforementioned issues, the main contributions of our paper include 
\begin{itemize} 
\item  A  band-stop  weight parametrization  that achieves a {\it joint} training of GCN topology and weights (see section \ref{section4.1}). This  parametrization relies on shared latent weights that reduce the number of training parameters    of the pruned GCNs. 
\item A KLD (Kullback Leibler Divergence) based regularizer that constrains the latent weights to fit an {\it a priori} distribution, and this  allows  implementing any targeted pruning rate  {\it almost} exactly (see section \ref{section4.2}). 
\item A multiple magnitude pruning that  obtains optimal  GCNs  at any targeted  pruning rate thanks to the band-stop parametrization and  the KLD regularizer. The latter defines a continuum of  {\it weight aggregates}  associated to GCNs with increasing pruning rates.  These {\it weight aggregates} allow generalizing across unseen pruning rates without retraining (see section \ref{section4.3}). 
\item Extensive experiments conducted on the challenging task of skeleton-based recognition corroborate all these findings and show the outperformance of our method against the related work (see section \ref{section5}). 
\end{itemize} 

\section{Related work}    
We  review and discuss subsequently  the related work in variational  pruning  and  skeleton-based recognition, and the limitations that motivate our contributions.\\ 

\noindent {\bf  Variational  Pruning.}  The general recipe of  variational pruning consists in learning both  weights and  binary masks that capture  topology of  pruned networks. This is achieved by minimizing a loss that combines (via a mixing hyperparameter) a classification error and a regularizer which controls the sparsity of the resulting masks \cite{DBLP:conf/iccv/LiuLSHYZ17,REFWen,REFICLR}. However, these methods are powerless to implement any given targeted pruning rate (cost) without overtrying multiple settings of the mixing hyperparameters. Alternative  methods explicitly model the cost, using $\ell_0$-based criteria   \cite{REFICLR,REFDrop}, in order to minimize the discrepancy between the observed cost and the targeted one.  Nonetheless, the underlying optimization problems are highly combinatorial and existing solutions usually rely on sampling heuristics or relaxation, such as $\ell_1$/$\ell_2$-based,  entropy,  etc.  \cite{REFGordon,REFCarreira,refref74,refref75}; the latter  promote sparsity, but are powerless to implement any given target cost exactly, and also result into overpruning effects leading to disconnected subnetworks, with weak generalization, especially at very high pruning regimes.  Besides, most of the existing solutions, including magnitude pruning \cite{DBLP:journals/corr/HanMD15}, decouple the training of network topology (masks) from weights, and this makes the learning of pruned networks clearly suboptimal.\\

 \noindent {\bf Skeleton-based Recognition.}  With the emergence of sensors, including Intel RealSense \cite{keselman2017} and Microsoft Kinect \cite{Zhang2017},  interest in skeleton-based  recognition is increasingly growing \cite{Cao2017}.  Hand-gesture and action recognition are two neighboring tasks  which have initially been tackled using RGB~\cite{Liu2018,refref18,wang2013directed,yuan2012mid,wang2014bags},  depth~\cite{refref39,Wang2018d},  shape/normals  \cite{refref40,refref41,Yun2012,Ji2014,Li2015a,refref59,sahbi2007kernel,sahbi2004kernel} and also skeleton-based  techniques \cite{Wang2018c}.  In particular,  early skeleton-based methods are  based on modeling human motions using handcrafted features \cite{Xia2012,Yang2014},    dynamic time warping~\cite{Vemulapalli2014}, temporal information \cite{refref61,refref11} as well as   temporal pyramids  \cite{Zhua2016,REF6}.   With the resurgence of deep learning,  all these methods have been quickly overtaken  by 2D/3D Convolutional Neural Networks (CNNs)  \cite{refref10,REF3,mazari2019deep}   that capture  global skeleton posture  together with local joint motion \cite{REF2,REF5},   by Recurrent Neural Networks (RNNs)  which capture motion dynamics \cite{Zhua2016,REF1,ke2017,ref32,Liu2018,ref53,Du2015,ref51,ref31,ref36,ref44,Zhang2017b,Lee2017,Liu2016,DeepGRU,Zhang2017,GCALSTM}, and manifold learning  \cite{Huangcc2017,ref23,Nguyen2019,Liu2021,RiemannianManifoldTraject}  as well as attention-based networks  \cite{REF7,REF2,REF9,Song2017}.    
With the recent emergence of GCNs,  the latter have been increasingly used  in skeleton-based  recognition  \cite{Huangcc2017,ref20,Lib2018,Yanc2018,Wen2019,Shi2018,Nguyen2019,Li2019,Jiang2020}.  These models explicitly capture,  with a better interpretability,   spatial and temporal attention among skeleton joints \cite{Li2018}.  However,  on  tasks involving large input graphs (such skeleton-based recognition),  GCNs become computationally inefficient and require lightweight design techniques.

  \section{Proposed Method}\label{section4.1}
 GCNs  correspond  to  convolutional blocks  with two layers; the first one aggregates signals (through  adjacency matrices)  while the second layer achieves convolution by multiplying the resulting aggregates with filters.    Learning  multiple adjacency (also referred to as attention) matrices  allows us to capture different contexts and graph topologies when achieving aggregation and convolution.    Stacking aggregation and convolutional layers, with multiple   attention matrices,   makes GCNs accurate but heavy.  We propose,  in what follows, a method that makes our networks lightweight and still effective. \\
 We formally subsume a given GCN as a multi-layered neural network $g_\theta$  whose weights are defined as $\theta =  \left\{\WW^1,\dots, \WW^L \right\}$, with $L$ being its depth,  $\WW^\ell \in \mathbb{R}^{d_{\ell-1} \times d_{\ell}}$ its $\ell^\textrm{th}$  layer weight tensor, and $d_\ell$ the dimension of $\ell$. The output of a  layer  $\ell$ is 
$ \mathbf{\phi}^{\ell} = f_\ell({\WW^\ell}^\top \  \mathbf{\phi}^{\ell-1})$, $\ell \in \{2,\dots,L\}$,  with  $f_\ell$ being an activation function.  

We consider a  {\it parametrization}, related to magnitude pruning, that allows defining  both the topology of the pruned networks and their weights.  This parametrization corresponds to the Hadamard product of  a weight tensor and a function applied entry-wise to the same tensor as
\begin{eqnarray}\label{eq2} 
\WW^\ell = \hat{\WW}^\ell \odot \psi(\hat{\WW}^\ell),
\end{eqnarray}

\begin{figure*}[h]
\centering
\resizebox{1.\columnwidth}{!}{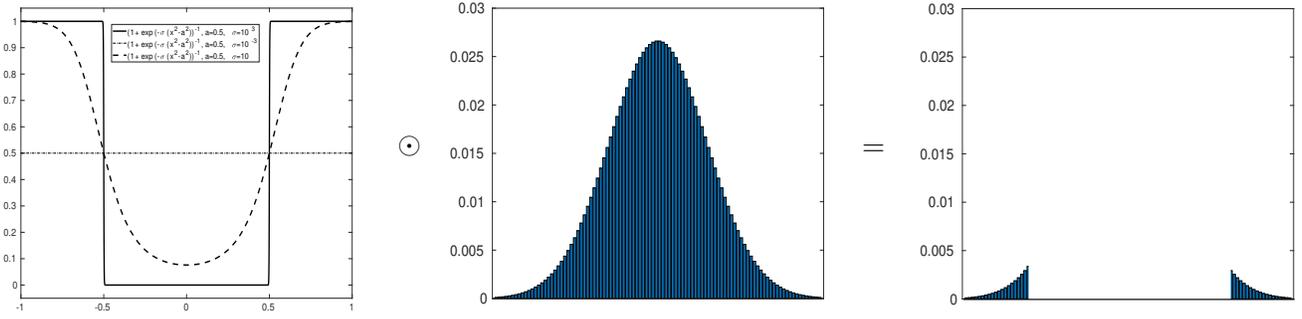}
\caption{This figure shows a Band-stop function $\psi_{a,\sigma}$ applied to  a given (gaussian) weight distribution. Depending on  $a$, only weights with  large magnitudes are kept (Better to zoom the file).}\label{tab2000}
\end{figure*}

\noindent here $\hat{\WW}^\ell$ is a latent tensor and $\psi(\hat{\WW}^\ell)$ is a continuous relaxation of the binary masks  which enforces the prior that smallest weights should be removed from the network.   A  possible choice, used in practice,   is $\psi_{a,\sigma}(\hat{\w})=(1+\sigma \exp (a^2-\hat{\w}^2))^{-1}$ being  $\sigma$  a scaling factor and ``$a$'' threshold.   As illustrated  in Fig.~\ref{tab2000},  $\sigma$ controls the smoothness of $\psi_{a,\sigma}$ around the support $\Omega \subseteq \mathbb{R}$ of the latent weights. This makes it possible to  implement  an annealed thresholding function that removes  connections in a smooth and differentiable manner as training of the latent parameters evolves.   In other words,  the asymptotic behavior of $\psi_{a,\sigma}$   --- that allows defining  the topology of the pruned networks  --- is obtained as training reaches the final  epochs. 
\subsection{SRMP: Single-Rate Magnitude Pruning}\label{section4.2}

The  parameterization described above  --- despite  being effective (as shown   later in  experiments) ---   does not allow to implement any a priori fixed pruning rate as the dynamic of the  weights $\{\hat{\WW}^\ell\}_\ell$ is not known beforehand.   Hence,  pruning rates could only be observed a posteriori or implemented after training using,  for instance,   a two stage process (namely magnitude pruning + fine-tuning).  In order to implement any targeted pruning rate in one  training step,  the distribution of latent weights is constrained to fit an arbitrary fixed  probability distribution,  so one may set  $a$ in $\psi_{a,\sigma}$ and hence  implement  the a priori fixed  pruning rate.  Let $\hat{W} \in \Omega$ be   a random variable standing for the latent weights in $g_\theta$; $\hat{W}$ is assumed taken  from any arbitrary distribution $P$ (uniform, gaussian, laplace, etc).  Setting  appropriately  $P$ (not only)  makes it possible to  implement  any targeted pruning rate,  but  also acts as a regularizer  that  controls the dynamic of the learned weights and thereby the generalization of   $g_\theta$ as shown  subsequently and also observed in experiments.\\

\noindent Let   $Q$ be  the observed distribution of  $\{\hat{\WW}^\ell\}_\ell$, and $P$ the targeted (fixed) one,  our objective  is to reduce the difference  between $P$ and $Q$ using a Kullback-Leibler Divergence (KLD) criterion   $D_{KL}(P||Q)$.   Notice  that the analytic form of  $D_{KL}(P||Q)$ is known on the widely used PDFs (probability density functions),   whilst for any arbitrary PDF,  the exact form is not always known.    Hence, we consider  instead an empirical (discrete)  variant of this criterion  and also  $P$,  $Q$;  here $P$ is  again the  targeted distribution  (such as gaussian,  laplace and uniform)  while the observed (and also differentiable) one $Q$ is based on a relaxed histogram estimation.  Let $\{q_1,\dots,q_K\}$ be a $K$-bin quantization of $\Omega$ ($K=100$ in practice),  the k-th entry of $Q$ is given by 
\begin{equation}\label{eq3}  
Q(\hat{W}=q_k) \propto \sum_{\ell=1}^{L-1} \sum_{i=1}^{n_\ell} \sum_{j=1}^{n_{\ell+1}} \exp\bigg\{-(\hat{\W}^\ell_{i,j}-q_k)^2/\beta_k^2\bigg\}, 
\end{equation}
\noindent here $\beta_k$ is a scaling coefficient  whose value  controls the smoothness of the exponential; high $\beta_k$ results  into flat (oversmoothed)  histograms  while a sufficiently (but not very) small $\beta_k$ allows obtaining surrogate histograms similar to the actual distribution of $Q$. In practice,  $\beta_k$ is fixed  to $(q_{k+1}-q_k)/2$; with this $\beta_k$,  the partition function of $Q$ --- i.e.,  $\sum_{k=1}^K Q(\hat{W}=q_{k})$ ---  reaches almost one in practice,   so  one may  replace $\propto$ (in Eq.~\ref{eq3})  with  equality.\\

\noindent Considering $F_{\hat{W}}(a)=P(\hat{W}\leq a)$ as  the cumulative distribution function (CDF) of $P(\hat{W})$.  For  any fixed  pruning rate $r$,   one may set  the threshold $a$ of  $\psi_{a,\sigma}$ as
\begin{equation}\label{eq333}   
a(r) = F_{\hat{W}}^{-1}(r),
 \end{equation}
which corresponds to the quantile that defines the pruning threshold on the targeted distribution $P$ (and also on the observed distribution $Q$ thanks to KLD).  This guarantees that only a fraction $r$ of the total weights are removed when applying the band-stop  in Eq.~\ref{eq2}.  It's worth noticing that   the quantile at any given $r$,   can either be analytically derived on the widely used PDFs  or empirically evaluated on discrete random variables. \\

 \begin{figure*}[h]
\centering
\resizebox{1.\columnwidth}{!}{
 \hspace{-0.3cm} \includegraphics[width=0.33\linewidth]{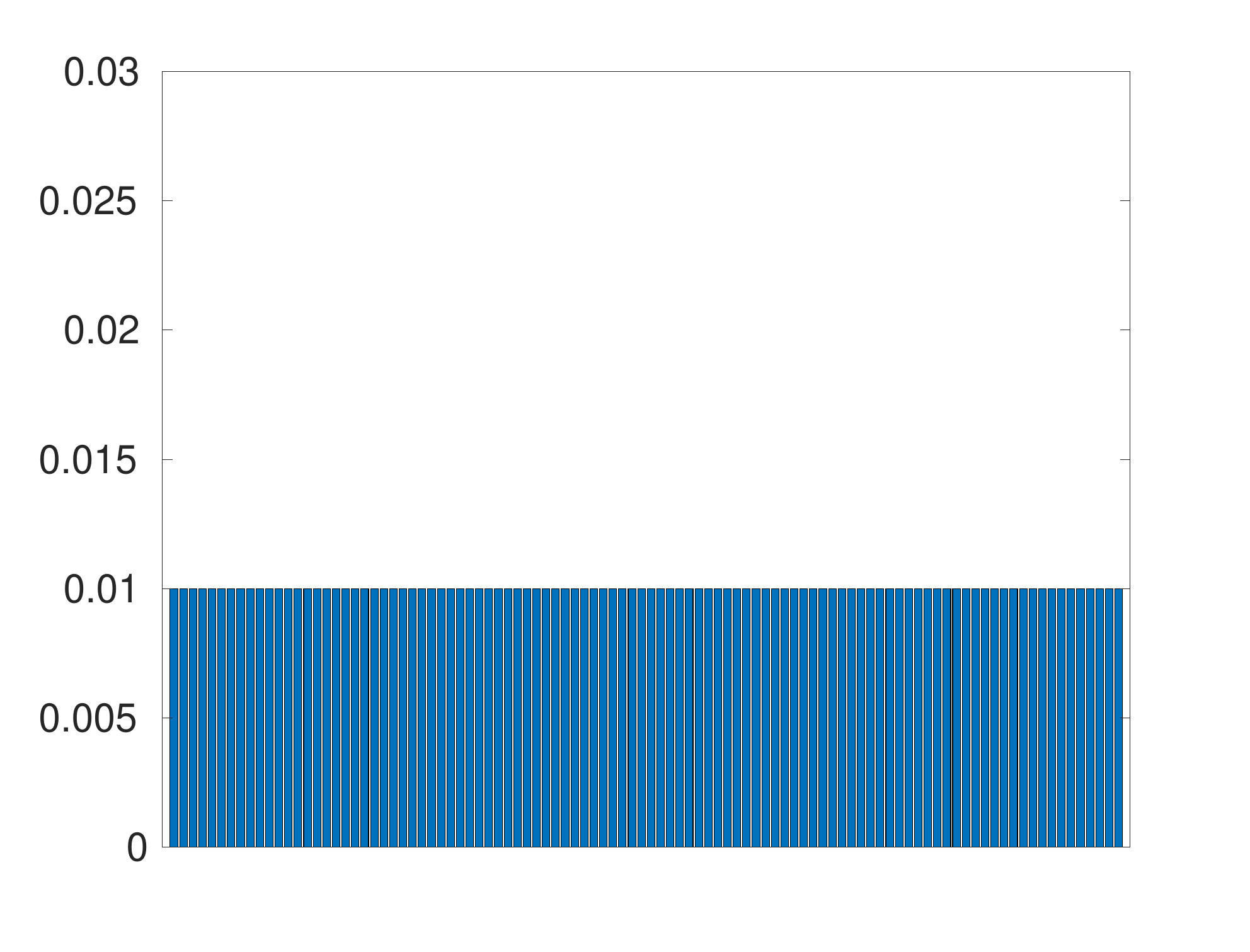} 
  \hspace{-0.3cm} \includegraphics[width=0.33\linewidth]{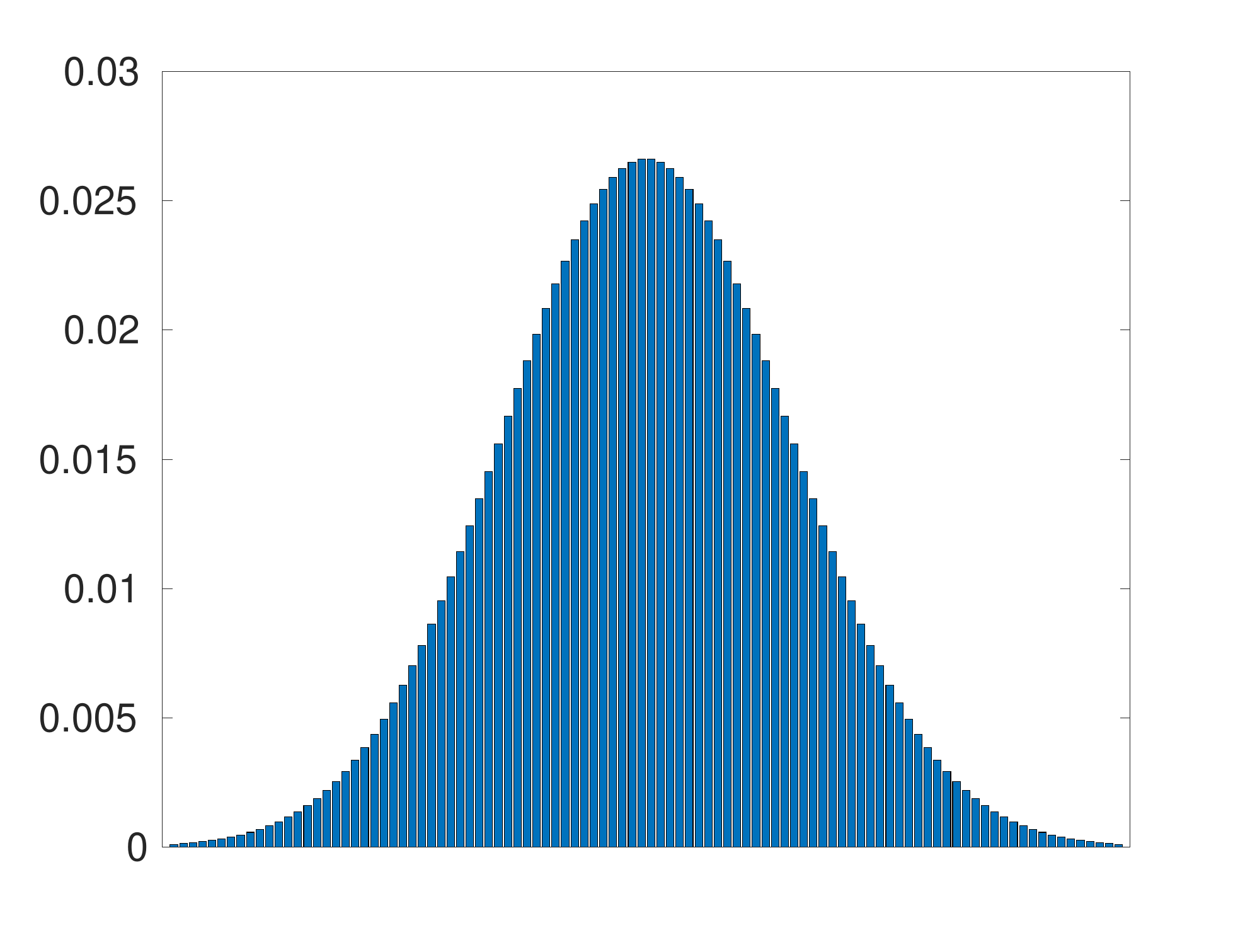}\hspace{-0.3cm} \includegraphics[width=0.33\linewidth]{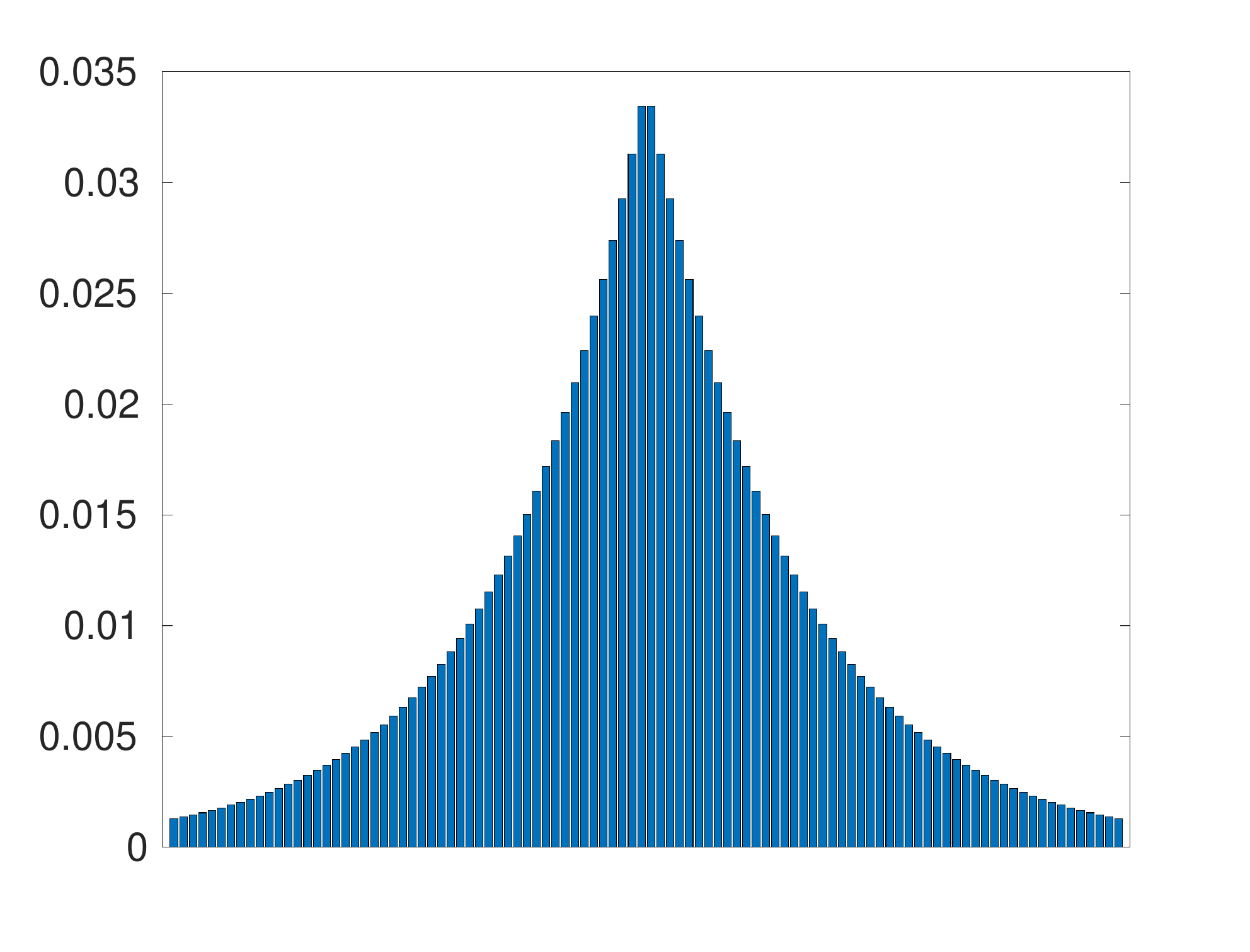} \hspace{-0.3cm}  \includegraphics[width=0.33\linewidth]{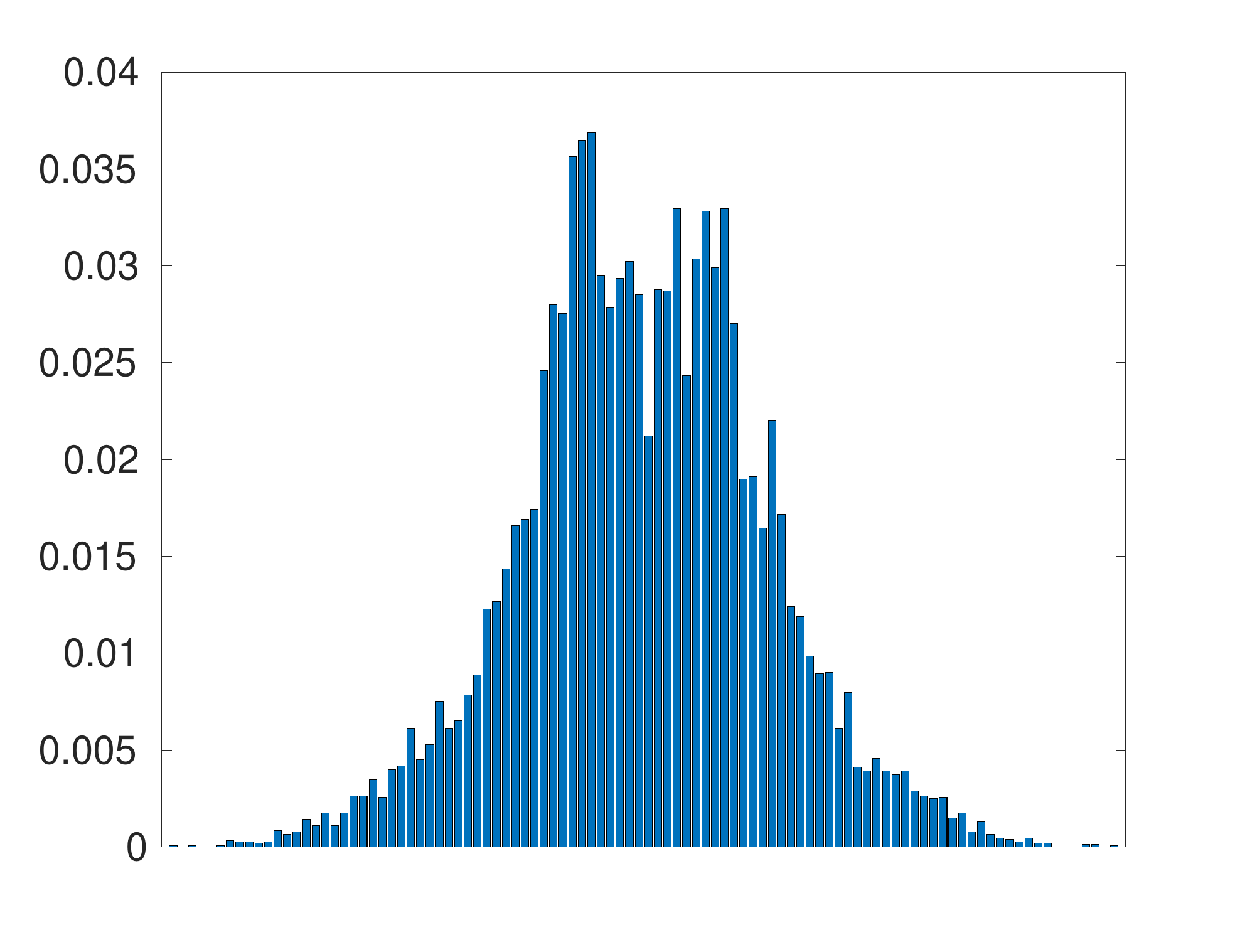}}
\caption{From left to right:  the  3 figures correspond to targeted (uniform, gaussian and laplace) distributions,  and the 4th  figure shows the actual weight distribution of the unpruned GCN which resembles to gaussian/laplacian.}\label{tab21}
\end{figure*}

\noindent Following  the aforementioned budget implementation,  pruning is achieved using an  objective function  that mixes  a cross-entropy term ${\cal L}_e$,   and the KLD criterion $D_{KL}$  resulting into
\begin{equation}\label{eq34} 
  \begin{array}{ll}
  \displaystyle  \min_{\{\hat{\WW}^\ell\}_\ell}  \displaystyle {\cal L}_e\big(\{\hat{\WW}^\ell \odot  \psi_{a,\sigma}(\hat{\WW}^\ell)\}_\ell\big)  \ + \ \lambda \ D_{KL}(P||Q), 
\end{array}
    \end{equation}
    \noindent   
 again  KLD controls weight distribution and it  guarantees  the targeted pruning budget depending on $a$ in $\psi_{a,\sigma}$ as shown in Eq.~\ref{eq333}.  In the above objective function,   $\lambda$ is overestimated (to $10$ in practice)  in order to make  Eq.~\ref{eq34} focusing   on the implementation of  the budget.  As training  reaches its final epochs,    $D_{KL}$ reaches  its minimum and the gradient of the global objective function becomes dominated by the gradient of ${\cal L}_e$, and this allows improving  further classification performances.

\subsection{MRMP: Multi-Rate Magnitude  Pruning}\label{section4.3}

The aforementioned formulation is already effective (see later experiments),  however, it requires rerunning a complete  optimization, for any update of the pruning rate which is time and memory demanding.  In what follows, we introduce a framework that allows training GCN instances with {\it shared latent weights}  which  achieve optimal performances at multiple pruning rates without retraining. \\ 
\noindent The guiding principle of our method relies on sharing the latent weights $\{\hat{\WW}^\ell\}_\ell$ through multiple GCN instances defined by the hyperparameters $\{a(r)\}_r$ in $\psi_{a,\sigma}$.  This makes it possible to reduce not only training time (as a unique training session is necessary for all the pruning rates) but also the memory footprint  as the number of latent weights remains unchanged.  Furthermore, training multiple GCN instances  on top of shared  latent parameters makes each instance as a proxy task to the other GCN learning tasks, and this turns out to improve generalization as shown in experiments.  Last but not least, this also allows obtaining optimal pruned networks (at unseen pruning rates) instantaneously without retraining.  In order to achieve these goals, we propose an updated loss as

\begin{equation}\label{eq35} 
 \hspace{-0.45cm} \begin{array}{ll}
  \displaystyle  \min_{\{\hat{\WW}^\ell\}_\ell}  \displaystyle \sum_r {\cal L}_e\big(\{\hat{\WW}^\ell \odot \psi_{a(r),\sigma}(\hat{\WW}^\ell)\}_\ell\big)  \ + \ \lambda \ D_{KL}(P||Q), 
\end{array}
    \end{equation}
  here the right-hand side term remains unchanged and it again seeks to constrain  the latent weights to fit a targeted distribution.  In contrast,  the left-hand side (cross entropy) term,  is evaluated through multiple pruning rates using the shared latent weights; hence only  $\psi_{a(r),\sigma}$ intervenes in order to prune connections according to the  targeted  rates.  Once the optimization achieved,   GCN instances may be obtained at any fixed pruning rates (including unseen ones) by multiplying each weight tensor with the binary mask tensor as $\{\hat{\WW}^\ell \odot \psi_{a,\sigma}(\hat{\WW}^\ell)\}_\ell$ here $a$ is again obtained using Eq.~\ref{eq333} and thanks to the KLD criterion in Eq.\ref{eq35}. It's worth noticing that the computational complexity of the above formulation is highly efficient (compared to running multiple independent instances of pruning); indeed,  the gradient of the KLD term is exactly the same while all the gradients of the  left-hand side terms (w.r.t. the GCN output) have similar analytic forms and their evaluation for different $r$ (either during forward and backward steps of backpropagation) could be batched and efficiently vectorized.   In practice, we observe a slight overhead of MRMP against SRMP (Single Rate Magnitude Pruning), in forward and backward steps, even when one hundred pruning rates ($r$) are jointly considered for  MRMP. 
  
       \begin{figure*}[tbp]
\center
\includegraphics[width=0.3\linewidth]{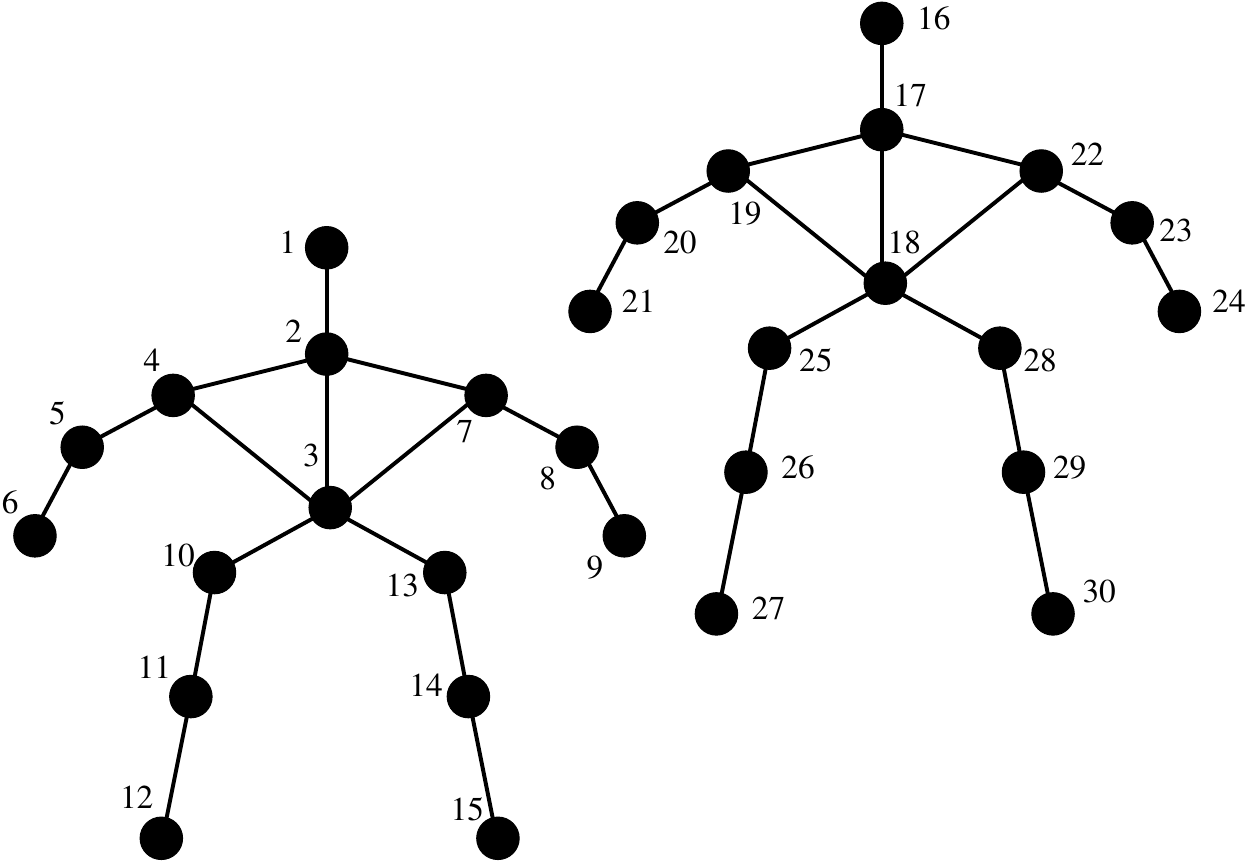}  \hspace{1cm}
\includegraphics[width=0.18\linewidth]{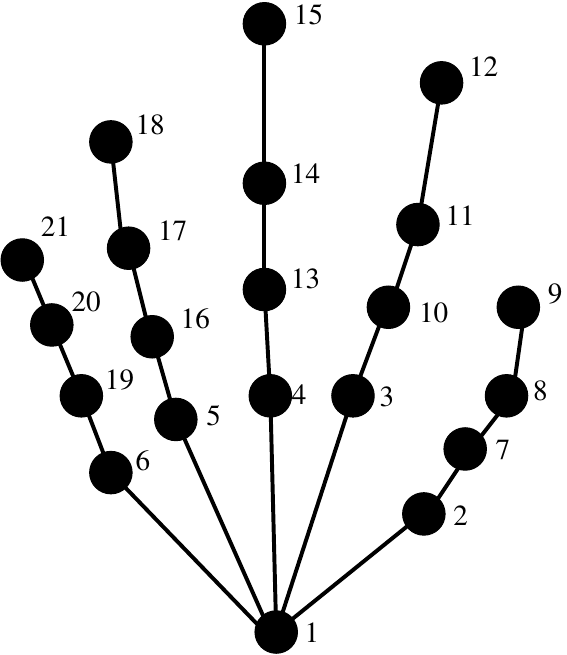}  \hspace{1cm} \scalebox{0.18}{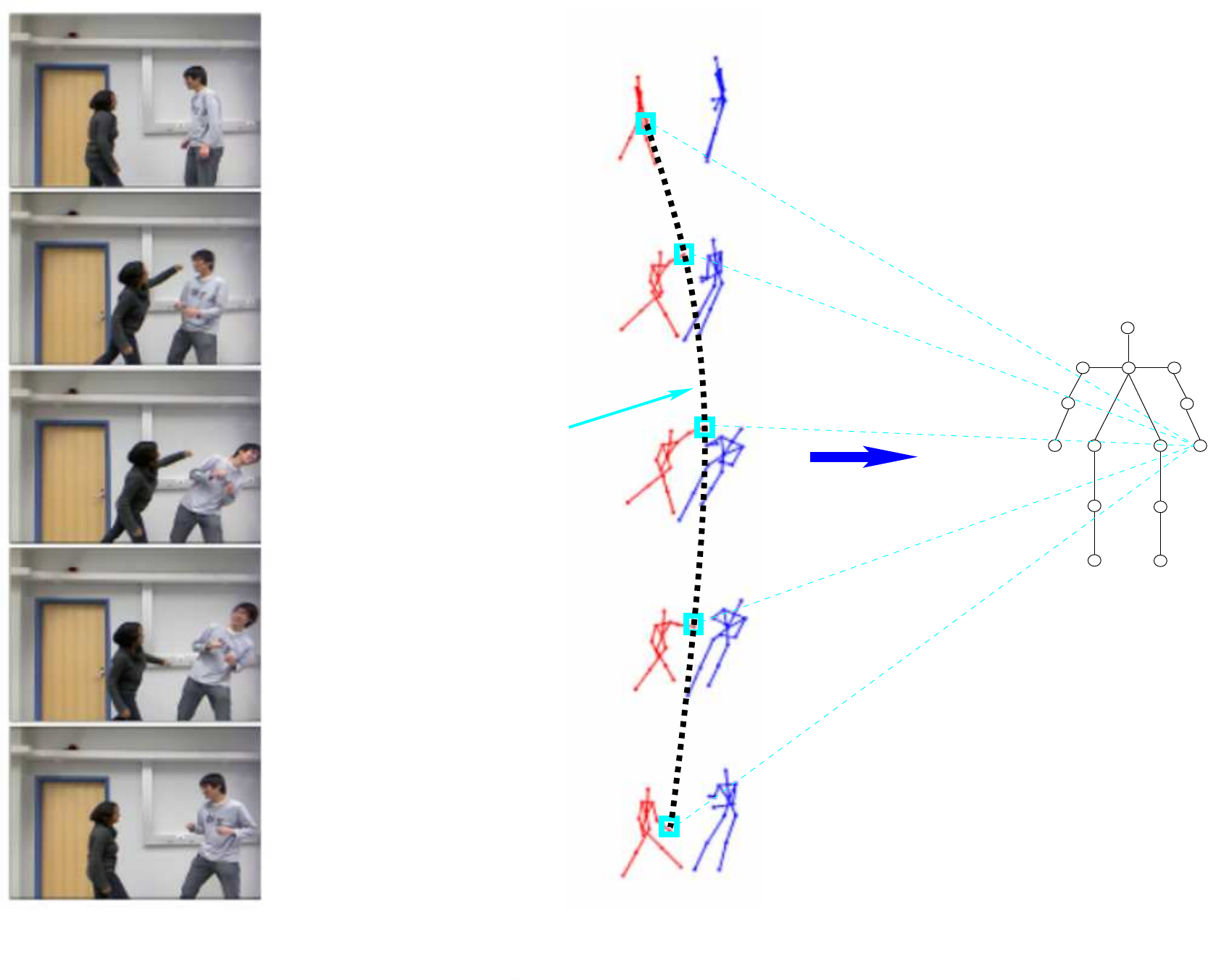}
\caption{This figure shows original skeletons (left) on the SBU and (middle) the FPHA datasets. (Right) this figure shows the whole keypoint  tracking and description process.}
\label{fig:A2}
\end{figure*}

 \begin{table}[ht]
\begin{center}
\resizebox{0.55\columnwidth}{!}{
\begin{tabular}{cc|c}
{\bf Method}      &   & {\bf Accuracy (\%)}\\
\hline 
  Raw Position \cite{Yun2012} & $ \ $   & 49.7   \\ 
  Joint feature \cite{Ji2014}  & $ \ $   & 86.9   \\
  CHARM \cite{Li2015a}       & $ \ $    & 86.9   \\
 \hline  
H-RNN \cite{Du2015}         & $ \ $    & 80.4   \\ 
ST-LSTM \cite{Liu2016}      & $ \ $    & 88.6    \\ 
Co-occurrence-LSTM \cite{Zhua2016} & $ \ $  & 90.4  \\ 
STA-LSTM  \cite{Song2017}     & $ \ $   & 91.5  \\ 
ST-LSTM + Trust Gate \cite{Liu2016} & $ \ $  & 93.3 \\
VA-LSTM \cite{Zhang2017}      & $ \ $  & 97.6  \\
 GCA-LSTM \cite{GCALSTM}                    &   $ \ $      &  94.9     \\ 
  \hline
Riemannian manifold. traj~\cite{RiemannianManifoldTraject} &  $ \ $  & 93.7 \\
DeepGRU  \cite{DeepGRU}        &    $ \ $   &    95.7    \\
RHCN + ACSC + STUFE \cite{Jiang2020} & $ \ $   & 98.7 \\ 
  \hline
\hline 
  Our baseline GCN &              &        98.4      
\end{tabular}}
 \end{center} 
  \caption{Comparison of our baseline GCN against related work on the SBU database.}\label{compare}
 \end{table} 
 
\section{Experiments}\label{section5}
 In this section, we evaluate the performance of our baseline and pruned GCNs on the task of skeleton-based recognition using two challenging skeleton datasets; SBU Interaction~\cite{Yun2012} and First-Person Hand Action (FPHA)~\cite{Garcia2018}.   The goal is to study the performance of our lightweight GCN design and its comparison against staple pruning techniques as well as the related work. \\

 \noindent {\bf Dataset description.} SBU is an interaction dataset acquired (under relatively well controlled conditions) using the Microsoft Kinect sensor; it includes in total 282 moving skeleton sequences (performed by two interacting individuals) belonging to 8 categories:  ``approaching'', ``departing'', ``pushing'', ``kicking'', ``punching'', ``exchanging objects'', ``hugging'', and ``hand shaking''.   Each pair of interacting individuals corresponds to two 15 joint  skeletons and each joint is encoded with a sequence of its 3D coordinates across video frames. In this dataset, we consider the same evaluation protocol as the one suggested in the original dataset release~\cite{Yun2012} (i.e., train-test split). \\ 
 The FPHA dataset includes 1175 skeletons  belonging to 45 action categories which are performed by 6 different individuals in 3 scenarios. In contrast to SBU, action categories are highly variable with inter and intra subject variability including style, speed, scale and viewpoint. Each skeleton includes 21 hand joints and each joint is again encoded with a sequence of its 3D coordinates across video frames. We evaluate the performance of our method using the 1:1 setting proposed in~\cite{Garcia2018} with 600 action sequences for training and 575 for testing. In all these experiments, we report the average accuracy over all the classes of actions.\\

  \noindent {\bf Skeleton normalization.} Let $S^t=\{p_1^t,\dots, p_n^t\}$ denote the 3D skeleton coordinates at frame $t$. Without a loss of generality, we consider a particular order so that $p_1^t$,  $p_2^t$ and $p_3^t$  correspond to three reference joints (e.g., neck, left shoulder and right shoulder); as shown in Fig.~\ref{fig:A2}, this corresponds to  joints 2, 4 and 7 for SBU and 1, 3 and 5 for FPHA. As the relative distance between these 3 joints is    stable w.r.t. any motion, these 3 joints are used in order to estimate the rigid motion (similarity transformation) for skeleton normalization; see also \cite{Meshry2016}. Each  graph sequence is  processed in order to normalize its 3D coordinates using a similarity transformation; the translation parameters ${\bf t}=(t_x,t_y,t_z)$  of this transformation correspond to the shift that makes the reference point $(p_2^0+p_3^0)/2$ coincide with the origin while the rotation parameters $({\bf \theta}_x,{\bf \theta}_y,{\bf \theta}_z)$ are chosen in order to make the plane formed by  $p_1^0$,  $p_2^0$ and $p_3^0$ coplanar with the x-y plane and the vector $p_2^0-p_3^0$ colinear with the x-axis. Finally, the scaling $\gamma$ of this similarity is chosen to make the $\|p_2^0-p_3^0\|_2$ constant through all the action instances. Hence, each normalized joint  is transformed  as $\hat{p}_i^t= \gamma ({p}_i^t-{\bf t})R_x({\bf \theta}_x) R_y({\bf \theta}_y) R_z({\bf \theta}_z)$ with $R_x$, $R_y$, $R_z$ being rotation matrices along the $x$, $y$ and $z$ axes respectively. \\

 \noindent {\bf Input graphs.}  Considering a sequence of normalized skeletons $\{S^t\}_t$, each joint sequence $\{\hat{p}_j^t\}_t$ in these skeletons defines a labeled trajectory  through successive frames (see Fig.~\ref{fig:A2}-right).   Given a finite collection of trajectories,  we consider the input graph $\G = (\V,\E)$ where each node $v_j \in \V$ corresponds to the labeled trajectory $\{\hat{p}_j^t\}_t$  and an edge $(v_j, v_i) \in  \E$ exists between two nodes iff the underlying trajectories are spatially neighbors. Each trajectory (i.e., node in $\G$) is processed using {\it temporal chunking}: first, the total duration of a  sequence (video) is split into $M$ equally-sized temporal chunks ($M=4$ in practice), then the normalized joint  coordinates  $\{\hat{p}_j^t\}_t$  of  the trajectory $v_j$ are assigned to the $M$ chunks (depending on their time stamps) prior to concatenate the averages of these chunks; this produces the description of $v_j$ (again denoted as $\phi(v_j) \in \mathbb{R}^{s}$ with $s=3 \times M$) and $\{\phi(v_j)\}_j$  constitutes the raw signal of nodes in a given sequence. Note that two trajectories $v_j$ and $v_i$,  with similar joint coordinates but arranged differently in time, will be considered as very different when using temporal chunking.   Note that temporal chunking produces discriminant raw descriptions that preserve the temporal structure of trajectories while being {\it frame-rate} and {\it duration} agnostic.\\

 \begin{table}[ht]
 \begin{center}
\resizebox{0.7\columnwidth}{!}{
\begin{tabular}{ccccc}
{\bf Method} & {\bf Color} & {\bf Depth} & {\bf Pose} & { \bf Accuracy (\%)}\\
\hline
  2-stream-color \cite{refref10}   & \cmark  &  \xmark  & \xmark  &  61.56 \\
 2-stream-flow \cite{refref10}     & \cmark  &  \xmark  & \xmark  &  69.91 \\  
 2-stream-all \cite{refref10}      & \cmark  & \xmark   & \xmark  &  75.30 \\
\hline 
HOG2-dep \cite{refref39}        & \xmark  & \cmark   & \xmark  &  59.83 \\    
HOG2-dep+pose \cite{refref39}   & \xmark  & \cmark   & \cmark  &  66.78 \\ 
HON4D \cite{refref40}               & \xmark  & \cmark   & \xmark  &  70.61 \\ 
Novel View \cite{refref41}          & \xmark  & \cmark   & \xmark  &  69.21  \\ 
\hline
1-layer LSTM \cite{Zhua2016}        & \xmark  & \xmark   & \cmark  &  78.73 \\
2-layer LSTM \cite{Zhua2016}        & \xmark  & \xmark   & \cmark  &  80.14 \\ 
\hline 
Moving Pose \cite{refref59}         & \xmark  & \xmark   & \cmark  &  56.34 \\ 
Lie Group \cite{Vemulapalli2014}    & \xmark  & \xmark   & \cmark  &  82.69 \\ 
HBRNN \cite{Du2015}                & \xmark  & \xmark   & \cmark  &  77.40 \\ 
Gram Matrix \cite{refref61}         & \xmark  & \xmark   & \cmark  &  85.39 \\ 
TF    \cite{refref11}               & \xmark  & \xmark   & \cmark  &  80.69 \\  
\hline 
JOULE-color \cite{refref18}         & \cmark  & \xmark   & \xmark  &  66.78 \\ 
JOULE-depth \cite{refref18}         & \xmark  & \cmark   & \xmark  &  60.17 \\ 
JOULE-pose \cite{refref18}         & \xmark  & \xmark   & \cmark  &  74.60 \\ 
JOULE-all \cite{refref18}           & \cmark  & \cmark   & \cmark  &  78.78 \\ 
\hline 
Huang et al. \cite{Huangcc2017}     & \xmark  & \xmark   & \cmark  &  84.35 \\ 
Huang et al. \cite{ref23}           & \xmark  & \xmark   & \cmark  &  77.57 \\  
\hline 
HAN  \cite{Liu2021}   & \xmark  & \xmark   & \cmark & 85.74 \\
  \hline
Our  baseline GCN                   & \xmark  & \xmark   & \cmark  &  86.43                                                 
\end{tabular}}
\caption{Comparison of our baseline GCN against related work on the FPHA database.}\label{compare2}
\end{center}
 \end{table}

 \begin{figure}[ht]
\center
 \includegraphics[width=0.6\linewidth]{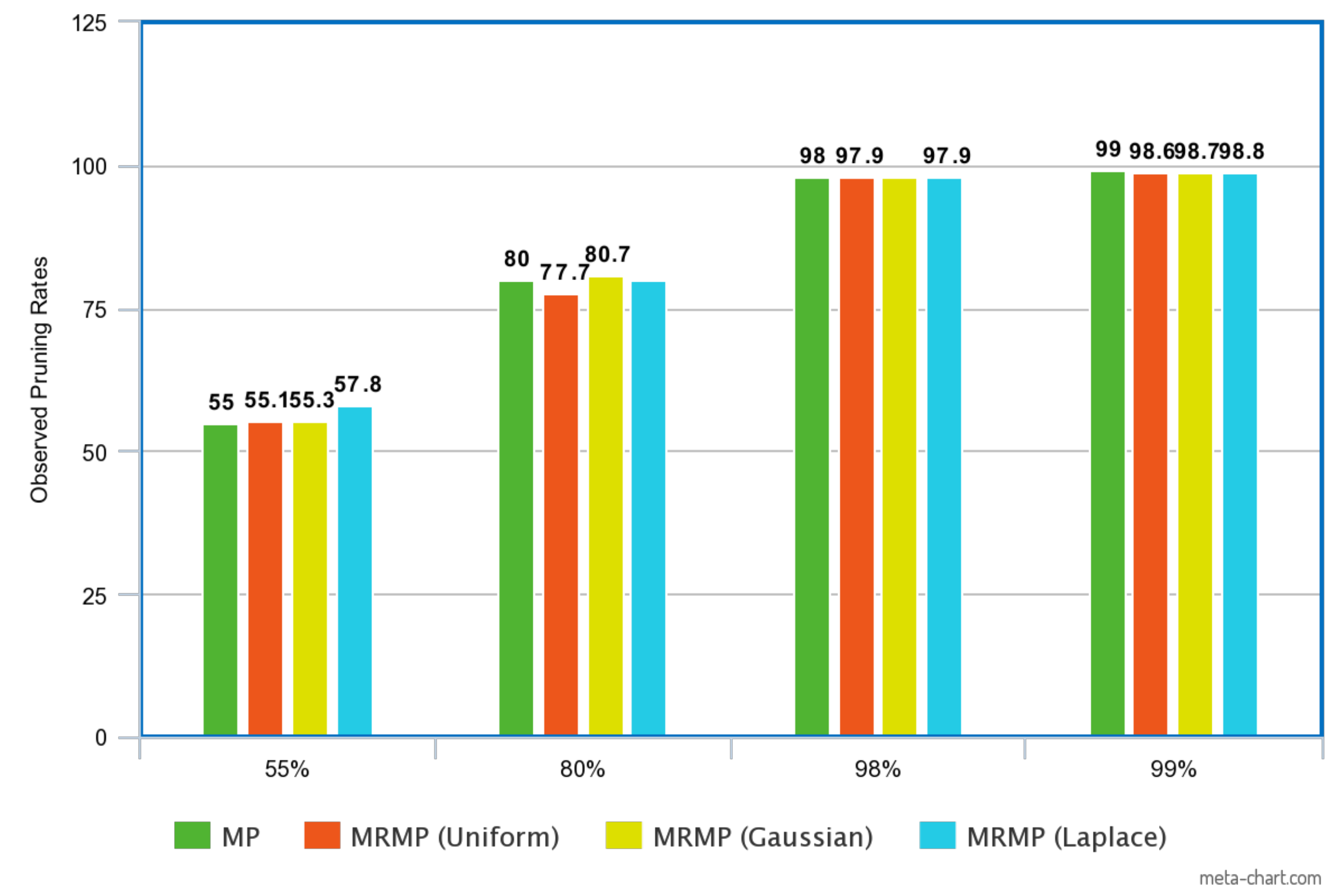} \includegraphics[width=0.7\linewidth]{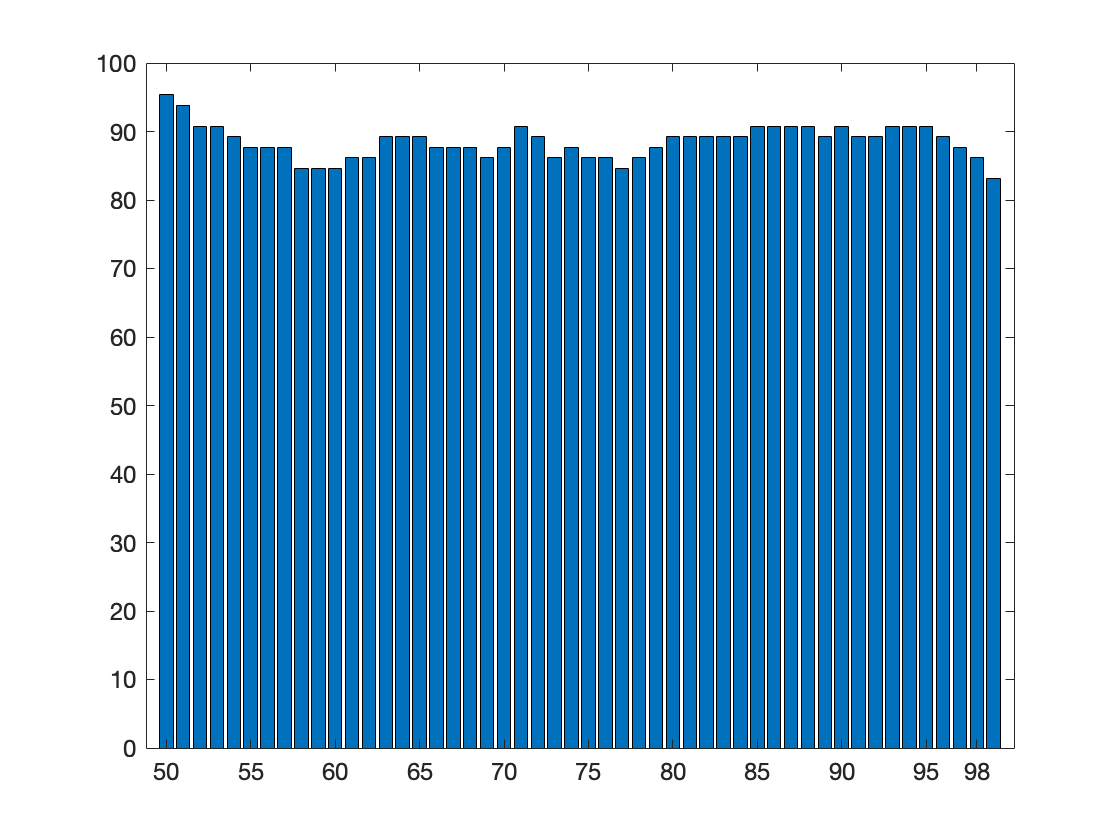}
\caption{Performances on the SBU dataset: (Top)  Fixed and observed pruning rates when different PDFs are used in the KLD regularizer. (Bottom) Performances for different (seen and unseen) pruning rates of MRMP; again seen pruning rates (during training) correspond to 50, 55, 60, 65, 70, 75, 80, 85, 90, 95 and 98\% while unseen ones correspond to all the remaining pruning rates in $[50,100[$.  Better to zoom the file.}
\label{fig:A3}
\end{figure}
 \begin{table}[h]
 \begin{center}
\resizebox{0.7\columnwidth}{!}{
  \begin{tabular}{ccc}    
   \rotatebox{0}{Pruning rates}  &     \rotatebox{0}{Accuracy (\%)}   & \rotatebox{0}{Observation}  \\
 \hline
  \hline
    0\%    &    \bf98.40   & Baseline GCN\\
 
                                70\% &  93.84 & Band-stop Weight Param.\\

    \hline
  \multirow{7}{*}{\rotatebox{0}{55\%}}     & 89.23  &  MP   \\
                                                              &   83.07   & SRMP: MP+KLD (Uniform) \ \ \ \ \ \ \ \ \          \\   
                                                              &   \bf93.84   & MRMP: MP+KLD (Uniform)+MR          \\   
                                                             &   81.53    & SRMP: MP+KLD (Gaussian)  \ \ \ \ \ \ \ \ \         \\
                                                        &   \bf90.76    & MRMP: MP+KLD (Gaussian)+MR         \\
                                                          &   80.00  & SRMP: MP+KLD (Laplace) \ \ \ \ \ \ \ \ \        \\
                                                          &   \bf89.23  & MRMP: MP+KLD (Laplace)+MR        \\

    \hline
  \multirow{7}{*}{\rotatebox{0}{80\%}}      & 87.69  & MP     \\
                                          &    84.61   & SRMP: MP+KLD (Uniform) \ \ \ \ \ \ \ \ \       \\
                                      &    \bf89.23   & MRMP: MP+KLD (Uniform)+MR        \\
                                                    &     78.46    & SRMP: MP+KLD (Gaussian)  \ \ \ \ \ \ \ \ \      \\
                                       &     \bf89.23    & MRMP: MP+KLD (Gaussian)+MR       \\
                                                           &  80.00  & SRMP: MP+KLD (Laplace)   \ \ \ \ \ \ \ \ \          \\
                                                &  \bf90.76  & MRMP: MP+KLD (Laplace)+MR            \\

 \hline  \multirow{7}{*}{\rotatebox{0}{98\%}}     &   69.23    & MP  \\
                                                              &   78.46     & SRMP:  MP+KLD (Uniform) \ \ \ \ \ \ \ \ \     \\
                                                         &   \bf86.15     & MRMP: MP+KLD (Uniform)+MR     \\
                                                               & 47.69 &  SRMP: MP+KLD (Gaussian)  \ \ \ \ \ \ \ \ \      \\
                                                & \bf86.15 &  MRMP: MP+KLD (Gaussian)+MR        \\
                                                            &  60.00  & SRMP: MP+KLD (Laplace)   \ \ \ \ \ \ \ \ \       \\
                                                    &    \bf80.00  & MRMP: MP+KLD (Laplace)+MR          \\

    \hline \hline
  \multicolumn{3}{c}{Comparative (regularization-based) pruning}   \\                              
    \hline 
      \multirow{4}{*}{\rotatebox{0}{98\%}}                                      &    55.38 & MP+$\ell_0$-reg. \\
                                                         &    73.84 & MP+$\ell_1$-reg. \\                                                                                                                                                                      
                                 &    61.53 & MP+Entropy-reg. \\ 
                                &   75.38 & MP+Cost-aware-reg.

  \end{tabular}}
\end{center}
\caption{Detailed performances and ablation study on SBU,  for different  pruning rates,     PDFs used for KLD and also MR;  MR stands for Multi-Rate pruning. }\label{table21}
\end{table}

\begin{table}[ht]
 \begin{center}
\resizebox{0.7\columnwidth}{!}{
  \begin{tabular}{ccc}    
   \rotatebox{0}{Pruning rates}    &      \rotatebox{0}{Accuracy (\%)}   & \rotatebox{0}{Observation}  \\
 \hline
  \hline
    0\%               &    \bf86.43  & Baseline GCN\\
 
                                50\% &  85.56 & Band-stop Weight Param.\\

    \hline
  \multirow{7}{*}{\rotatebox{0}{55\%}}     & 87.82  &  MP  \\
                                                           &   87.82   & SRMP: MP+KLD (Uniform)  \ \ \ \ \ \ \ \ \            \\
                                                             &  \textcolor{black}{\bf88.92}   & MRMP: MP+KLD (Uniform)+MR           \\
                                                           &  88.52     & SRMP: MP+KLD (Gaussian) \ \ \ \ \ \ \ \ \           \\
                                                           &    \textcolor{black}{\bf89.58}     & MRMP: MP+KLD (Gaussian)+MR         \\
                                                          &  87.65  & SRMP: MP+KLD (Laplace)   \ \ \ \ \ \ \ \ \        \\
                                                            &  \textcolor{black}{\bf88.48}  & MRMP: MP+KLD (Laplace)+MR        \\

    \hline
  \multirow{7}{*}{\rotatebox{0}{80\%}}      & 86.78  & MP     \\
                                                           &    85.91     & SRMP: MP+KLD (Uniform)  \ \ \ \ \ \ \ \ \        \\
                                                         &    \textcolor{black}{\bf86.43}     & MRMP: MP+KLD (Uniform)+MR        \\ 
                                                           &   87.47     & SRMP: MP+KLD (Gaussian) \ \ \ \ \ \ \ \ \       \\
                                                      &    \textcolor{black}{\bf88.93}    & MRMP: MP+KLD (Gaussian)+MR      \\
                                                        &  86.95  & SRMP: MP+KLD (Laplace)   \ \ \ \ \ \ \ \ \         \\
                                                        &  \textcolor{black}{\bf87.89}  & MRMP: MP+KLD (Laplace)+MR         \\
 
 \hline  \multirow{7}{*}{\rotatebox{0}{98\%}}       &   60.34    & MP  \\
                                                       &   70.26      & SRMP: MP+KLD (Uniform)  \ \ \ \ \ \ \ \ \      \\
                                                    &     \textcolor{black}{\bf71.18}      & MRMP: MP+KLD (Uniform)+MR      \\
                                                          & 70.60  &  SRMP: MP+KLD (Gaussian)    \ \ \ \ \ \ \ \ \      \\
                                                         &  \textcolor{black}{\bf74.73}  &  MRMP: MP+KLD (Gaussian)+MR       \\
                                                             &  70.80   & SRMP: MP+KLD (Laplace)  \ \ \ \ \ \ \ \ \         \\
                                                             &   \textcolor{black}{\bf72.97}   & MRMP: MP+KLD (Laplace)+MR          \\
 
\hline 
\hline   \multicolumn{3}{c}{Comparative (regularization-based) pruning}   \\                              
    \hline 
           \multirow{4}{*}{\rotatebox{0}{98\%}}                                   &   64.69 & MP+$\ell_0$-reg. \\
                                                    &    70.78 &  MP+$\ell_1$-reg. \\  
                             &   67.47& MP+Entropy-reg. \\  
                             &   69.91 & MP+Cost-aware-reg.

  \end{tabular}}
\end{center}

\caption{Same caption as  tab~\ref{table21} but for FPHA dataset.}\label{table22}
\end{table}

 \noindent {\bf Implementation details \& baseline GCNs.}  We trained the GCN  networks end-to-end using the Adam optimizer \cite{Adam2014} for 2,700 epochs  with a batch size equal to $200$ for SBU and $600$ for FPHA, a momentum of $0.9$ and a global learning rate (denoted as $\nu(t)$)  inversely proportional to the speed of change of the loss used to train our networks; when this speed increases (resp. decreases),   $\nu(t)$  decreases as $\nu(t) \leftarrow \nu(t-1) \times 0.99$ (resp. increases as $\nu(t) \leftarrow \nu(t-1) \slash 0.99$). All these experiments are run on a GeForce GTX 1070 GPU device (with 8 GB memory) and neither dropout nor data augmentation are used.  The architecture of our baseline GCN includes an attention layer of 1 head on SBU (resp. 16 heads on FPHA) applied to skeleton graphs whose nodes are encoded with 8-channels (resp. 32 for FPHA), followed by a convolutional layer of 32 filters for SBU (resp. 128 filters for FPHA), and a dense fully connected layer and a softmax layer. The initial network for SBU is not very heavy, its number of parameters does not exceed 15,320, and this makes its pruning challenging as many connections will be isolated (not contributing in the evaluation of the network output). In contrast, the initial network for FPHA is relatively heavy (for a GCN) and its number of parameters reaches 2 millions.  As shown in tables.~\ref{compare} and \ref{compare2}, both GCNs are accurate compared to the related work on the SBU/FPHA benchmarks. Considering these GCN baselines, our goal is to make them highly lightweight while making their accuracy as high as possible.\\
 
\noindent{\bf Performances, Comparison \& Ablation.}  Tables~\ref{table21}-\ref{table22}  and  Fig.~\ref{fig:A3} show a comparison and an ablation study of our method both on SBU and FPHA datasets.  First,  from the results in Fig.\ref{fig:A3}-top, we see the alignment between the targeted pruning rates and the observed ones when using the formulation in Eq. \ref{eq34} for different PDFs; the quantile functions of the gaussian and laplace PDFs allow implementing {\it fine-steps} of the targeted pruning rates $r$  particularly when $r$ is large.   In contrast, the quantile functions of the gaussian and laplace PDFs are coarse around mid $r$ values ($55\%$).  Second, according to tables \ref{table21}-\ref{table22},  when training is achieved with only the band-stop weight parametrization (i.e., $\lambda=0$ in  Eq. \ref{eq34}), performances are close  to the initial heavy GCNs (particularly on FPHA), with less parameters\footnote{Pruning rate does not exceed 70\% and no control on this rate is achievable.} as this produces a regularization effect similar to \cite{dropconnect2013}.  Third, we observe a positive impact when the KLD term (in  Eq. \ref{eq34}) is used both with single and multi-rate magnitude pruning (resp. SRMP and MRMP) with an extra-advantage of MRMP against SRMP; again MP+KLD in tables~\ref{table21}-\ref{table22}  corresponds to SRMP and  MP+KLD+MR refers to MRMP.   Extra comparison of KLD against other regularizers shows the substantial gain of our method.  Indeed,  KLD is  compared against different alternatives plugged in Eq. \ref{eq34} instead of KLD,  namely $\ell_0$ \cite{REFICLR},  $\ell_1$ \cite{refref74},  entropy \cite{refref75} and $\ell_2$-based cost  \cite{REFLemaire}.   From the observed results,   the impact of  KLD  is substantial for different PDFs and for  equivalent pruning rate (namely 98\%).  Note that when alternative regularizers are used, multiple settings (trials) of the underlying hyperparameter $\lambda$ (in Eq. \ref{eq34}) are considered prior to reach the  targeted pruning rate, and this makes the whole training and pruning process overwhelming. While cost-aware regularization makes training more tractable, its downside resides in the observed collapse of trained masks; this is a well known effect that affects performances  at high pruning rates.   Finally,  Fig.\ref{fig:A3}-bottom  shows  the generalization performance of MRMP from seen to unseen pruning rates. In these results,   only a few pruning rates  are used for MRMP (namely $50$,  $55$,  $60$,  $65$,  $70$,  $75$,  $80$,  $85$,  $90$,  $95$ and $98\%$) and  all the remaining rates in $[50, 100[$ are used for instantaneous pruning without retraining.  We observe stable and high performances {\it from seen to unseen} pruning rates and this shows that MRMP  is able to extrapolate highly accurate GCNs (even) on unseen pruning rates.

\section{Conclusion}
We introduce in this paper a novel lightweight GCN design based on multi-rate magnitude pruning.  Our method allows training multiple network instances simultaneously, on top of shared latent weights, at different pruning rates and extrapolating GCNs at unseen rates without retraining their weights. Experiments conducted on the challenging task of skeleton-based recognition, using two different standard datasets,   show a significant gain of our method against the related work.   As a perspective,  we are currently investigating the extension of the current approach to other deep neural architectures  and applications.


\begin{thebibliography}{100}
\bibitem{aaaaa9}  H. Sahbi.  A particular Gaussian mixture model for clustering and its application to image retrieval.   Soft Computing 12 (7), 667-676
\end{thebibliography}
\end{document}